\documentclass{article}

\usepackage[preprint]{neurips_2026}

\usepackage[utf8]{inputenc} % allow utf-8 input
\usepackage[T1]{fontenc}    % use 8-bit T1 fonts
\usepackage{hyperref}       % hyperlinks
\usepackage{url}            % simple URL typesetting
\usepackage{booktabs}       % professional-quality tables
\usepackage{amsfonts}       % blackboard math symbols
\usepackage{nicefrac}       % compact symbols for 1/2, etc.
\usepackage{microtype}      % microtypography
\usepackage{xcolor}         % colors

\usepackage{array}
\newcolumntype{H}{>{\setbox0=\hbox\bgroup}c<{\egroup}@{}}
\usepackage{mystyle}
\usepackage{authblk}

\DeclareAcronym{api}{
	short = API,
	long = {Application Program Interface}
}

\DeclareAcronym{awgn}{
	short = AWGN,
	long = {additive white Gaussian noise}
}

\DeclareAcronym{vae}{
	short = VAE,
	long = {Variational AutoEncoder}
}

\DeclareAcronym{bert}{
	short = BERT,
	long = {Bidirectional Encoder Representations from Transformers}
}

\DeclareAcronym{roberta}{
	short = RoBERTa,
	long = {Robustly optimized BERT approach}
}

\DeclareAcronym{ast}{
	short = AST,
	long = {Abstract Syntax Tree}
}

\DeclareAcronym{bpe}{
	short = BPE,
	long = {Byte-Pair Encoding}
}

\DeclareAcronym{cfg}{
	short = CFG,
	long = {Control Flow Graph}
}

\DeclareAcronym{dcg}{
	short = DCG,
	long = {Discounted Cumulative Gain}
}

\DeclareAcronym{gpt}{
	short = GPT,
	long = {Generative Pretrained Transformer}
}

\DeclareAcronym{ir}{
	short = IR,
	long = {Information Retrieval}
}

\DeclareAcronym{lstm}{
	short = LSTM,
	long = {Long Short-Term Memory}
}

\DeclareAcronym{clm}{
	short = CLM,
	long = {Casual Language Modeling}
}

\DeclareAcronym{mlm}{
	short = MLM,
	long = {Masked Language Modeling}
}

\DeclareAcronym{mem}{
	short = MEM,
	long = {Multimodal Embedding Model}
}

\DeclareAcronym{cp}{
	short = CP,
	long = {Continuous Pretraining}
}

\DeclareAcronym{if}{
	short = IF,
	long = {Intermediate Finetuning}
}

\DeclareAcronym{mmpf}{
	short = MMPF,
	long = {Massive Multitask Pre-Finetuning}
}

\DeclareAcronym{aif}{
	short = AIF,
	long = {Adaptive Intermediate Finetuning}
}

\DeclareAcronym{mrr}{
	short = MRR,
	long = {Mean Reciprocal Rank}
}

\DeclareAcronym{ndcg}{
	short = NDCG,
	long = {Normalized Discounted Cumulative Gain}
}

\DeclareAcronym{nlp}{
	short = NLP,
	long = {Natural Language Processing}
}

\DeclareAcronym{nlp_pt}{
	short = NLP\textsubscript{PT},
	long = {Next Line Prediction}
}

\DeclareAcronym{nmt}{
	short = NMT,
	long = {Neural Machine Translation}
}

\DeclareAcronym{nsp}{
	short = NSP,
	long = {Next Sentence Prediction}
}

\DeclareAcronym{rnn}{
	short = RNN,
	long = {Recurrent Neural Network}
}

\DeclareAcronym{cnn}{
	short = CNN,
	long = {Convolutional Neural Network}
}

\DeclareAcronym{tf-idf}{
	short = tf-idf,
	long = {term frequency–-inverse document frequency}
}

\DeclareAcronym{anova}{
	short = ANOVA,
	long = {ANalysis Of VAriance}
}

\DeclareAcronym{da}{
	short = DA,
	long = {Domain-Adaptive}
}

\DeclareAcronym{ta}{
	short = TA,
	long = {Task-Adaptive}
}

\DeclareAcronym{ma}{
	short = MA,
	long = {Multiphase Adaptive}
}

\DeclareAcronym{ca}{
	short = CA,
	long = {Concept Annotation}
}

\DeclareAcronym{ce}{
	short = CE,
	long = {Concept Extrapolation}
}

\DeclareAcronym{ci}{
	short = CI,
	long = {Concept Interpolation}
}

\DeclareAcronym{gru}{
	short = GRU,
	long = {Gated Recurrent Unit}
}

\DeclareAcronym{sota}{
	short = SOTA,
	long = {state-of-the-art}
}

\DeclareAcronym{lcs}{
	short = LCS,
	long = {Longest Common Sequences}
}

\title{Rethinking Weight Tying: Pseudo-Inverse Tying for LM Stable Training and Updates}

\author[1]{Jian Gu}
\author[1]{Aldeida Aleti}
\author[2]{Chunyang Chen}
\author[3]{Hongyu Zhang}
\affil[1]{Monash University, \texttt{jian.gu,aldeida.aleti@monash.edu}}
\affil[2]{Technical University of Munich, \texttt{chun-yang.chen@tum.de}}
\affil[3]{Chongqing University, \texttt{hyzhang@cqu.edu.cn}}

\begin{document}

\maketitle

\begin{abstract}
Weight tying is widely used in compact language models to reduce parameters by sharing the token table between the input embedding and the output projection. However, parameter sharing alone does not guarantee a stable token interface: during training, the correspondence between encoding tokens into hidden states and decoding hidden states into logits can drift, worsening optimization sensitivity and weakening explainability probes that rely on a meaningful vocabulary-space decoder. We propose Pseudo-Inverse Tying (PIT), which synchronizes embedding and unembedding as coupled projections of a shared latent token memory, guaranteeing a pseudo-inverse-consistent interface throughout training. PIT maintains an orthonormal shared memory, obtained by polar initialization from a source checkpoint for continued pretraining or by random orthonormal initialization for from-scratch pretraining, and introduces a learned symmetric positive definite hidden-space transform parameterized via a Cholesky factor. The output head applies this transform to hidden states before the vocabulary projection, while the embedding applies the inverse transform to token vectors using stable triangular solves, avoiding explicit pseudo-inverse recomputation and vocabulary-sized auxiliary parameters. Beyond improving training stability, PIT provides a cleaner substrate for logit-lens-style and vocabulary-space explainability probes by keeping the input and output token geometries synchronized. We evaluate PIT on on-device models spanning 256M-1.3B parameters. The results show that PIT improves continued-pretraining stability, enforces near-exact token-interface consistency across settings, and yields more predictable lightweight adaptation after continued pretraining, while from-scratch pretraining reveals a trade-off between strict interface consistency and unconstrained optimization.
\end{abstract}

% TLDR
% Pseudo-Inverse Tying (PIT) is a simple, low-cost diagonal-spectrum replacement for standard weight tying that keeps the embedding–unembedding interface well-conditioned, making on-device LM train and update more stably without side effects.

% the main figure shall be used in a future paper, discussing the changes of semantic bases in middle layers
% it shall be after the LLM proxy...
\section{Introduction}
\label{sec:introduction}

Compact language models are increasingly deployed in resource-constrained settings, such as on-device assistants, private enterprise endpoints, and edge applications, where latency, memory footprint, and deployment simplicity are first-class constraints. In this regime, practitioners often rely on parameter sharing and lightweight output heads to keep both training and inference practical. A widely adopted technique is \emph{weight tying}~\cite{inan2016tying,press2017using}, which reuses the input embedding as the output projection, typically via a transpose. Weight tying reduces parameters and often improves sample efficiency, making it especially attractive for small and medium-scale models.

Despite its practical success, standard weight tying only \emph{loosely} couples the two ends of the token interface. It enforces shared parameters, but it does not explicitly control the \emph{geometric compatibility} between encoding discrete tokens into continuous hidden states and decoding hidden states back into token logits. This distinction matters because the embedding-unembedding pair is the model's \emph{token interface}: a bottleneck through which all symbolic information enters and leaves the residual stream. During training, the effective encoding and decoding maps can become subtly misaligned, especially when optimization is sensitive or when the tokenizer introduces a large vocabulary. We refer to this phenomenon as \emph{token-interface drift}. In practice, drift can surface as unstable optimization, heightened sensitivity to learning-rate schedules, and brittle behavior under post-training modifications that assume a stable token-aligned geometry.

Token-interface stability is also a prerequisite for reliable explainability probes that interpret hidden states through the vocabulary space. Mechanistic interpretability tools such as the logit lens decode intermediate residual-stream states with the model's unembedding, implicitly treating the output head as a semantic coordinate system for layerwise analysis~\cite{nostalgebraist2020logitlens,elhage2021mathematical}. Related vocabulary-space probes, editing diagnostics, and adaptation analyses make a similar assumption: a hidden-state direction should have a stable token-level meaning when it is decoded. When the token interface drifts, the same hidden-space displacement can be read through a changing output geometry, making intermediate predictions, causal traces, and post-training updates harder to compare across layers or checkpoints.

This motivation aligns with the emerging view of \emph{intrinsic interpretability}, which aims to build transparency directly into model architectures and computations rather than relying only on post-hoc approximations. A recent survey organizes this direction around design paradigms such as functional transparency, concept alignment, representational decomposability, explicit modularization, and latent sparsity induction~\cite{gao2026towards}. PIT contributes to this agenda at the token-interface level. Instead of adding an external explainer after training, PIT makes the embedding and unembedding mutually consistent by construction, so vocabulary-level probes such as the logit lens operate on a more stable and geometrically aligned decoder. This does not by itself solve interpretability, but it removes a source of avoidable ambiguity from analyses that depend on vocabulary directions.

We propose Pseudo-Inverse Tying (PIT), a simple and principled mechanism that enforces a pseudo-inverse-consistent token interface throughout training. PIT synchronizes the embedding and unembedding by constructing them as coupled projections of a shared latent token memory: the two matrices are derived from the same underlying representation through an invertible coupling. This design guarantees that decoding acts as a left-inverse of encoding in the model's hidden space, so the token interface remains aligned rather than drifting during optimization. We implement the coupling with a learnable symmetric positive definite transform, parameterized via a Cholesky factor, which provides expressive cross-dimension mixing while remaining well-conditioned and independent of vocabulary size. To obtain a canonical shared memory, PIT uses thin polar decomposition for continued pretraining from a source checkpoint and random orthonormal initialization for from-scratch pretraining. It applies the coupling transform to hidden states in the output head and uses stable triangular solves to map token vectors through the inverse transform, avoiding repeated pseudo-inverse recomputation or vocabulary-sized auxiliary parameters.

We evaluate PIT on compact LMs spanning 256M-1.3B parameters across continued pretraining, from-scratch pretraining, and lightweight downstream adaptation. In continued pretraining, PIT improves training loss and perplexity for the diverse Granite and Qwen3 models we test while preserving comparable runtime. Across both continued and from-scratch settings, it enforces near-exact input-output semantic alignment by construction. The from-scratch and adaptation results are more nuanced: the strict interface constraint can trade off against unconstrained optimization, but it provides a clear stabilizing bias for continued-pretraining checkpoints and makes post-training behavior easier to diagnose.

To summarize, our contributions are as follows:
\begin{itemize}
    \item We identify \emph{token-interface drift} as a practical limitation of standard weight tying in compact and large-vocabulary language models, with implications for optimization stability, explainability probes, and post-training controllability.
    \item We propose \emph{Pseudo-Inverse Tying (PIT)}, a shared-memory parameterization that synchronizes the input embedding and output projection by construction.
    \item We present an efficient and deployment-friendly realization of PIT based on polar canonicalization and a learned positive-definite transform, avoiding repeated pseudo-inverse computation and vocabulary-sized auxiliary parameters.
    \item We evaluate PIT across model scales, tokenizers, and training regimes, showing improved continued-pretraining stability, near-exact token-interface consistency, and more predictable lightweight adaptation after continued pretraining.
\end{itemize}

\section{Related Work}
\label{sec:related_work}

\paragraph{Weight Tying and Token Interfaces}
Sharing the input embedding and the output projection (\emph{weight tying}) is a long-standing practice in neural language modeling, motivated by parameter efficiency and improved generalization~\cite{inan2016tying,press2017using}. Classic tying typically reuses the token embedding matrix as the output classifier (often via transpose), providing a simple coupling between token encoding and token decoding. However, standard tying does not explicitly enforce a stable \emph{token interface} during training: the geometry connecting input token vectors, hidden states, and output logits can drift under optimization, scaling, or post-training modifications. Pseudo-Inverse Tying (PIT) strengthens this coupling by constructing the embedding and unembedding as synchronized projections of a shared latent token memory, ensuring an explicitly pseudo-inverse-consistent interface throughout training.

\paragraph{LM Training and Updates}
A growing line of work studies how to update language models reliably, including targeted rewriting of factual associations and controlled downstream adaptation. Model editing methods such as ROME~\cite{meng2022locating} and its scalable extension MEMIT~\cite{meng2022mass} demonstrate that specific factual behaviors can be modified via structured weight updates, while highlighting practical difficulties such as interference, spillover, and sensitivity to representation drift. More broadly, many deployment-centric workflows (continual training, patching, and lightweight adaptation) rely on predictable mappings between hidden states and token logits to measure progress and validate updates. PIT complements these directions by stabilizing the token interface at the architectural level, reducing a source of training-time drift that can otherwise amplify the unpredictability of post-training updates.

\paragraph{LM Mechanistic Interpretability}
Mechanistic interpretability and representation analysis often decode intermediate activations back into vocabulary space to probe what a model ``knows'' at different depths. The \emph{logit lens} popularized layerwise decoding by applying the model's unembedding to intermediate residual-stream states~\cite{nostalgebraist2020logitlens}, and subsequent frameworks formalized transformer computations as compositions of interpretable components and circuits~\cite{elhage2021mathematical}. These analyses implicitly assume that the token interface remains meaningful and comparably scaled across layers and across training checkpoints. By enforcing a pseudo-inverse-consistent interface through synchronized projections, PIT improves the reliability of intermediate decoding and provides a cleaner substrate for layerwise semantic analyses.

\paragraph{Optimization with Orthogonality Constraints}
PIT uses an orthonormal shared token memory (a Stiefel-constrained matrix) so that pseudo-inverse consistency holds by construction. Enforcing orthogonality with low overhead is closely related to classical optimization on matrix manifolds~\cite{absil2008optimization} and practical feasible methods for orthogonality-constrained optimization~\cite{wen2013feasible}. In our setting, orthogonality is not an end goal; rather, it serves as a computationally convenient mechanism to maintain a well-defined shared memory and a mathematically guaranteed token interface across training and updates.

\section{Preliminaries}
\label{sec:prelim_pit}

\subsection{Semantic Property in Latent Space}

Next-token prediction creates a simple but revealing interface between discrete tokens and continuous hidden states.
A token is read in through $W_{\mathrm{in}}$, transformed layer by layer into a latent representation, and then converted back to a distribution over the vocabulary through $W_{\mathrm{out}}$.
This makes it natural to describe what the model ``means'' by a token using objects that live in latent space.

Recent work on LM semantics~\cite{gu2024vds,gu2025semantic,gu2025beyond} promotes a geometric viewpoint where each vocabulary label corresponds to a distinguished latent vector, so that tokens have explicit representatives inside the model.
On the input side, these representatives are already visible, since each row of $W_{\mathrm{in}}$ is how the model instantiates a token internally.
On the output side, the LM head defines how hidden states are interpreted as vocabulary scores, and we can place output labels into the same latent-space coordinates by mapping them back through the Moore-Penrose pseudoinverse.

Let $V$ be LM vocabulary size and $d$ the model dimension, with $V \gg d$.
We denote the input embedding matrix by $W_{\mathrm{in}} \in \mathbb{R}^{V \times d}$ and the unembedding matrix by $W_{\mathrm{out}} \in \mathbb{R}^{d \times V}$.
% For simplicity, we omit the bias term in the geometric discussion below, although it is of included in practice.
This yields two vocabulary-indexed collections of latent vectors
\begin{align}
B_{\mathrm{in}}
&= W_{\mathrm{in}}
\in \mathbb{R}^{V \times d},
\\
B_{\mathrm{out}}
&= W_{\mathrm{out}}^{+}
\in \mathbb{R}^{V \times d},
\end{align}
where $(\cdot)^{+}$ denotes the Moore-Penrose pseudoinverse.

\begin{figure}[ht]
    \centering
    \includegraphics[width=0.55\linewidth]{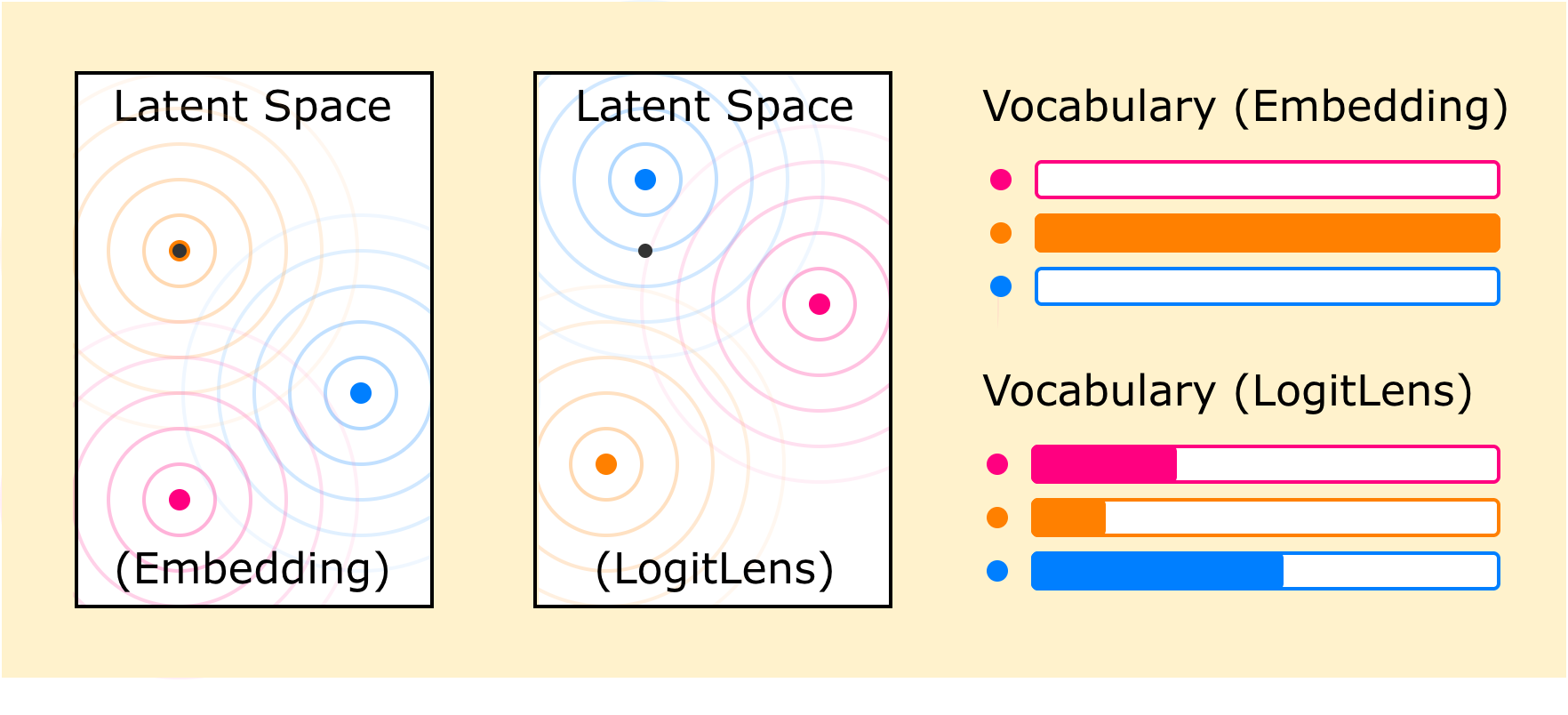}
    \caption{Two perspectives of token semantics in the residual stream: input-side ($B_{\mathrm{in}}$) vs.\ output-side ($B_{\mathrm{out}}$, Logit Lens).}
    \label{fig:perspective}
\end{figure}

Figure~\ref{fig:perspective} visualizes these collections as colored dots, one per vocabulary label, together with an arbitrary hidden state shown as a dark dot.
The left panel shows the input-side perspective induced by $B_{\mathrm{in}}$.
The right panel shows the output-side perspective induced by $B_{\mathrm{out}}$, which is essentially the same linear readout used by LogitLens when it probes intermediate hidden states.
When these two perspectives are not aligned, the same latent direction can correspond to different token meanings depending on whether we interpret it through the input interface or the output interface.

Given such vocabulary representatives, the scores of a hidden state can be understood geometrically as comparisons to all vocabulary anchors.
One can view the pre-softmax scores as induced by distances or similarities between the hidden state and the anchor set.
Numerically this recovers the standard matrix-multiplication logits, but the geometric language is helpful for understanding how emphasizing or suppressing a particular semantic meaning corresponds to moving a representation toward or away from specific anchors.
In Figure~\ref{fig:perspective}, the dashed connections from the dark dot to colored dots depict these comparisons.

This geometry also provides a lens on the forward pass itself.
Given a sequence $t_1, t_2, \ldots, t_n$, the model uses the last token $t_n$ as the medium through which it predicts $t_{n+1}$.
Let $f_0, f_1, \ldots, f_m$ denote the hidden states of this medium token across an $m$-layer transformer, where $f_0$ is the embedding lookup and $f_m$ is the final-layer representation used for prediction.
As the representation flows through layers, its nearest or most compatible vocabulary anchors may change, yielding a layer-by-layer evolution of its implied meaning.
We refer to this evolution as a semantic transition, and the sequence $\{f_\ell\}_{\ell=0}^m$ records a transition trace of how next-token prediction emerges through depth.

\subsection{Token-Interface Consistency and Weight Tying}

The geometric picture above depends not only on representations $f_\ell$, but also on the token interface that defines the vocabulary anchors themselves.
If the anchors used to read tokens in and the anchors used to read predictions out are synchronized, then token meanings are consistent across the model's input and output boundaries.
Algebraically, this synchronization corresponds to the input-side and output-side anchor sets agreeing, which can be expressed as
$W_{\mathrm{in}} \triangleq W_{\mathrm{out}}^{+}$.

\begin{figure}[ht]
    \centering
    \includegraphics[width=1.0\linewidth]{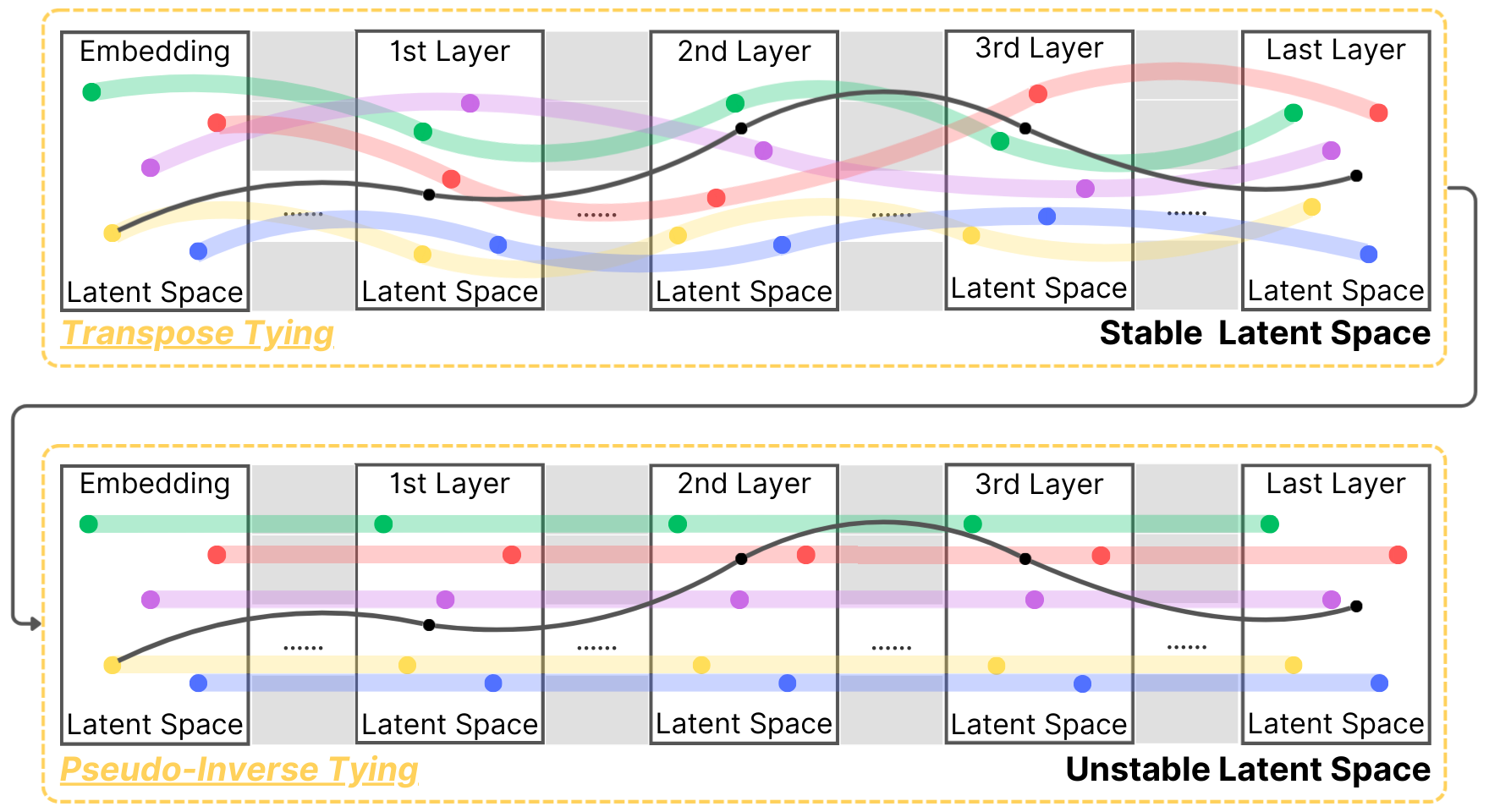}
    \caption{Layerwise semantic transition (transition trace) of the medium token during next-token prediction, under a stable vs.\ drifting token interface.}
    \label{fig:overview}
\end{figure}

Figure~\ref{fig:overview} illustrates this process as a trajectory of the medium token across layers.
By comparing each $f_\ell$ to the vocabulary anchors in the corresponding latent space, we can associate each layer with the token meaning that is most strongly supported at that depth.
When the latent semantics are stable, this transition is smooth and interpretable.
When the latent semantics drift, the same geometric neighborhood can correspond to different labels at different layers, making the transition appear erratic.

In practice, most language models parameterize $W_{\mathrm{in}}$ and $W_{\mathrm{out}}$ independently, or tie them by a transpose.
Transpose tying reduces parameters, but it does not in general enforce a pseudo-inverse relationship, and therefore does not guarantee that $B_{\mathrm{in}}$ and $B_{\mathrm{out}}$ stay aligned.
When this misalignment grows during pretraining, the model effectively uses one coordinate system to write tokens into the network and a slightly different one to read them out.
Figure~\ref{fig:overview} shows how such a mismatch manifests across depth.
Under transpose tying, the semantic geometry drifts from layer to layer, visualized as shifting color bands, and the medium token can appear to ``change its nearest meaning'' even when its trajectory is smooth in raw hidden-state space.
In contrast, when the input and output interfaces are synchronized, the anchor set remains stable across layers, and the semantic transition of the medium token is measured against a consistent vocabulary geometry.

A seemingly direct remedy would periodically recompute a pseudoinverse with SVD and retie the matrices, but doing so is computationally expensive and can be numerically brittle at scale.
Motivated by the token-interface viewpoint and the layerwise effects highlighted in Figure~\ref{fig:overview}, we instead seek to maintain input-output semantic synchronization throughout training by enforcing a structural consistency condition at the interface itself
\begin{equation}
W_{\mathrm{out}}\,E \approx I_d,
\label{eq:target_pinv_consistency}
\end{equation}
where $E\in\mathbb{R}^{V\times d}$ is the embedding and $W_{\mathrm{out}}\in\mathbb{R}^{d\times V}$ is the unembedding.
The key goal is to achieve Eq.~\eqref{eq:target_pinv_consistency} by construction rather than by a soft regularizer, while keeping the method stable and practical for large-scale pretraining.
This is the core motivation for Pseudo-Inverse Tying, which we introduce next.

\section{Approach: Pseudo-Inverse Tying}
\label{sec:approach}

We generalize standard weight tying into a \emph{shared-memory} view: both embedding and unembedding are derived from a \emph{single shared token memory} and two \emph{coupled} inverse transforms.
Concretely, we introduce a shared memory matrix $Z\in\mathbb{R}^{V\times d}$ and an invertible transform $T\in\mathbb{R}^{d\times d}$, and define
\begin{equation}
E \;=\; Z\,T^{-1},
\qquad
W_{\mathrm{out}} \;=\; T\,Z^\top.
\label{eq:spt_def}
\end{equation}
Eq.~\eqref{eq:spt_def} can be understood as \emph{pseudo-inverse tying via shared projection}: $Z$ stores token semantic memories as rows, while $T$ specifies a learnable change of metric that is applied to the input and output sides through $T^{-1}$ and $T$.

If $Z$ has orthonormal columns,
\begin{equation}
Z^\top Z \;=\; I_d,
\label{eq:stiefel_Z}
\end{equation}
then pseudo-inverse consistency holds exactly:
\begin{equation}
W_{\mathrm{out}}\,E
\;=\;
(TZ^\top)(ZT^{-1})
\;=\;
T\,(Z^\top Z)\,T^{-1}
\;=\;
I_d.
\label{eq:pinv_exact}
\end{equation}
Therefore, PIT reduces the problem of maintaining Eq.~\eqref{eq:target_pinv_consistency} to two structural requirements: $Z$ stays on the Stiefel manifold, and $T$ remains invertible. In this work, we choose $T$ to be symmetric positive definite (SPD).

\subsection{Polar Decomposition for Shared Memory}
\label{subsec:polar}

In practice, we start from either a source-checkpoint embedding or a random Gaussian initialization, each represented as an arbitrary tall matrix $\tilde{Z}\in\mathbb{R}^{V\times d}$.
To satisfy Eq.~\eqref{eq:stiefel_Z} in a numerically stable and canonical way, we use the thin polar decomposition:
\begin{equation}
\tilde{Z} \;=\; U\,H,
\qquad
U^\top U = I_d,
\qquad
H \succ 0.
\label{eq:polar}
\end{equation}
We then set the shared memory to the orthonormal factor:
\begin{equation}
Z \leftarrow U.
\label{eq:Z_from_polar}
\end{equation}
Polar decomposition is convenient because it separates an orthonormal basis from a symmetric positive definite scale/mixing component, aligning with our goal of disentangling shared token memory from the learnable metric.

We use two initialization regimes.
(i) In \emph{continued pretraining}, the Transformer parameters are initialized from an existing checkpoint and PIT is enabled for the continued-pretraining run. In this case, $Z$ is initialized from the checkpoint embedding via polar decomposition.
(ii) In \emph{from-scratch pretraining}, all model parameters are initialized randomly and PIT is enabled throughout training. In this case, $Z$ is initialized from a random Gaussian matrix followed by polar decomposition or thin QR.

\paragraph{Initialization for continued pretraining}
Let $E_0\in\mathbb{R}^{V\times d}$ be the embedding table from the source checkpoint.
We compute $E_0=U H$ and set $Z\leftarrow U$.
A simple practice is to initialize $T\leftarrow I$ and learn $T$ during training.

\paragraph{Initialization for from-scratch pretraining}
We sample $\tilde{Z}\sim\mathcal{N}(0,1)^{V\times d}$ and compute its polar or thin QR factorization to obtain an orthonormal $Z$ satisfying Eq.~\eqref{eq:stiefel_Z}.

\subsection{Full SPD Transform for Pseudo-Inverse Tying}
\label{subsec:learn_T}

\paragraph{Constraint on $T$}
We learn $T$ as a full \emph{symmetric positive definite} matrix:
\begin{equation}
T \succ 0.
\label{eq:T_spd}
\end{equation}
Choosing $T$ to be SPD ensures invertibility, improves numerical stability, and provides richer expressivity than diagonal or basis-diagonal scaling: a full SPD transform can represent both scaling and mixing through its eigenstructure.

\paragraph{Parameterization by Cholesky}
We parameterize $T$ via a Cholesky factor:
\begin{equation}
T \;=\; L L^\top,
\label{eq:chol_T}
\end{equation}
where $L$ is lower-triangular with strictly positive diagonal.
In implementation, we learn unconstrained entries of $L$ and enforce $L_{ii}>0$ using a softplus or exponential, optionally with a mild clamp for conditioning.

\paragraph{Efficient Forward Computation}
PIT does not require materializing $E$ and $W_{\mathrm{out}}$ as standalone parameters; instead, both are computed from $(Z,T)$.
Given hidden states $h\in\mathbb{R}^{BT\times d}$ (batch size $B$, sequence length $T$), logits are computed as
\begin{equation}
\mathrm{logits}(h)
\;=\;
h\,W_{\mathrm{out}}
\;=\;
h\,(T Z^\top).
\label{eq:logits_TZt}
\end{equation}
We implement Eq.~\eqref{eq:logits_TZt} as two matrix multiplications:
\begin{equation}
g \leftarrow h\,T \in \mathbb{R}^{BT\times d},
\qquad
\mathrm{logits} \leftarrow g\,Z^\top \in \mathbb{R}^{BT\times V}.
\label{eq:two_step_logits}
\end{equation}
The embedding lookup for token indices $t$ uses
\begin{equation}
e_t \;=\; z_t\,T^{-1},
\label{eq:embed_lookup}
\end{equation}
where $z_t$ is the $t$-th row of $Z$.
To avoid explicitly forming $T^{-1}$, we compute $e_t$ via Cholesky solves using $L$ from Eq.~\eqref{eq:chol_T}, which is more stable in mixed precision.

\section{Experiments}
\label{sec:experiments}

We evaluate PIT under continued pretraining, from-scratch pretraining, and lightweight post-training updates.
PIT constructs the embedding and unembedding as \emph{coupled projections} of a shared token memory using an orthonormal memory $Z\in\mathbb{R}^{V\times d}$ and a learned SPD transform $T\in\mathbb{R}^{d\times d}$:
\[
E=ZT^{-1},\qquad W_{\text{out}}=TZ^\top,
\]
which guarantees a pseudo-inverse-consistent token interface ($W_{\text{out}}E=I_d$) whenever $Z^\top Z=I_d$ and $T$ is invertible.
Our experiments are designed to answer three questions.
(i) Does PIT improve training stability and convergence at fixed compute.
(ii) Does it better preserve input-side vs.\ output-side semantics throughout training.
(iii) Does this translate into more predictable performance under lightweight post-training updates.

For pretraining, we report learning curves and fixed-budget training metrics.
For interface analysis, we report deterministic semantic-alignment diagnostics computed from trained checkpoints.

\subsection{Experimental Setups}
\label{sec:exp_overview}

\paragraph{Baselines}
We compare \emph{PIT} with conventional \emph{transpose tying}, shortened as \emph{TT}.

\paragraph{Models}
We evaluate two groups of models. The first group contains diverse checkpoints for continued pretraining, including Granite~\cite{Soule_Bergmann_2025} and Qwen3~\cite{yang2025qwen3}, each with base and instruct variants.
The second group contains scaled models for both continued and from-scratch pretraining: openly released \emph{Cerebras-GPT}~\cite{dey2023cerebras} models at 256M, 590M, and 1.3B parameters, and Llama-like models at 303M, 613M, and 1.2B parameters that we train from random initialization.
Compared with GPT-2~\cite{yenduri2023generative}, Cerebras-GPT better fits our controlled continued-pretraining setting because it uses tied embeddings, a no-bias LM head, and a consistent layer design across scales.
Together, these models span on-device to edge-server scales and allow us to test when structural token-interface constraints are most useful.

\subsection{Weight Tying for LM Continued Pretraining}

Table~\ref{tab:continued} summarizes continued-pretraining results for Granite and Qwen3 checkpoints.
Across all four checkpoints, PIT achieves lower final training loss and perplexity than TT under comparable runtime.
The improvement is largest for the Granite checkpoints, where PIT roughly halves training perplexity relative to TT, and remains visible for Qwen3 despite its larger vocabulary and different checkpoint family.
The runtime differences are small relative to the total run time: PIT is slightly faster for Granite4-350M-Base and Qwen3-0.6B-Base, while slightly slower for the instruct variants.
This pattern suggests that the added hidden-space transform is not the dominant cost; performance differences are primarily driven by optimization behavior rather than raw throughput.

\begin{table}[ht]
\centering
\small
\caption{Effects of weight tying on LM continued pretraining for diverse models.}
\label{tab:continued}

\resizebox{0.7\linewidth}{!}{

\begin{tabular}{llrrr}
  \toprule
  Model & Method & Training Loss $\downarrow$ & Training PPL $\downarrow$ & Runtime $\downarrow$ \\
  \midrule
  \multirow{2}{*}{\makecell{Granite4-350M-Base}} & TT & 3.482 & 32.517 & 2h 03m \\
                    & PIT & \tabh 2.760 & \tabh 15.796 & \tabh 2h 01m \\
  \midrule
  \multirow{2}{*}{\makecell{Granite4-350M-Instruct}} & TT & 3.527 & 34.011 & \tabh 2h 01m \\
                    & PIT & \tabh 2.636 & \tabh 13.964 & 2h 13m \\
  \midrule
  \multirow{2}{*}{\makecell{Qwen3-0.6B-Base}} & TT & 3.526 & 33.976 & 3h 50m \\
                    & PIT & \tabh 3.336 & \tabh 28.111 & \tabh 3h 47m \\
  \midrule
  \multirow{2}{*}{\makecell{Qwen3-0.6B-Instruct}} & TT & 3.539 & 34.416 & \tabh 3h 48m \\
                    & PIT & \tabh 3.237 & \tabh 25.469 & 3h 58m \\
  \bottomrule
  \end{tabular}

}
\end{table}

\paragraph{Analysis}
The continued-pretraining results support the central claim that stabilizing the token interface can reduce optimization friction after checkpoint initialization.
PIT improves loss more strongly for Granite than for Qwen3, which is plausible because the benefit depends on how much input-side and output-side token geometry in the source checkpoint has drifted before the continued-pretraining run.
The instruct checkpoints also benefit, indicating that the effect is not limited to base models.
At the same time, the gains should not be read as a universal throughput advantage: PIT's computational overhead is small but nonzero, so its practical value comes from better convergence and a more stable interface at approximately fixed wall-clock cost.

\subsection{Results on Loss Dynamics}

\begin{figure}[ht]
  \centering
  \includegraphics[width=\linewidth]{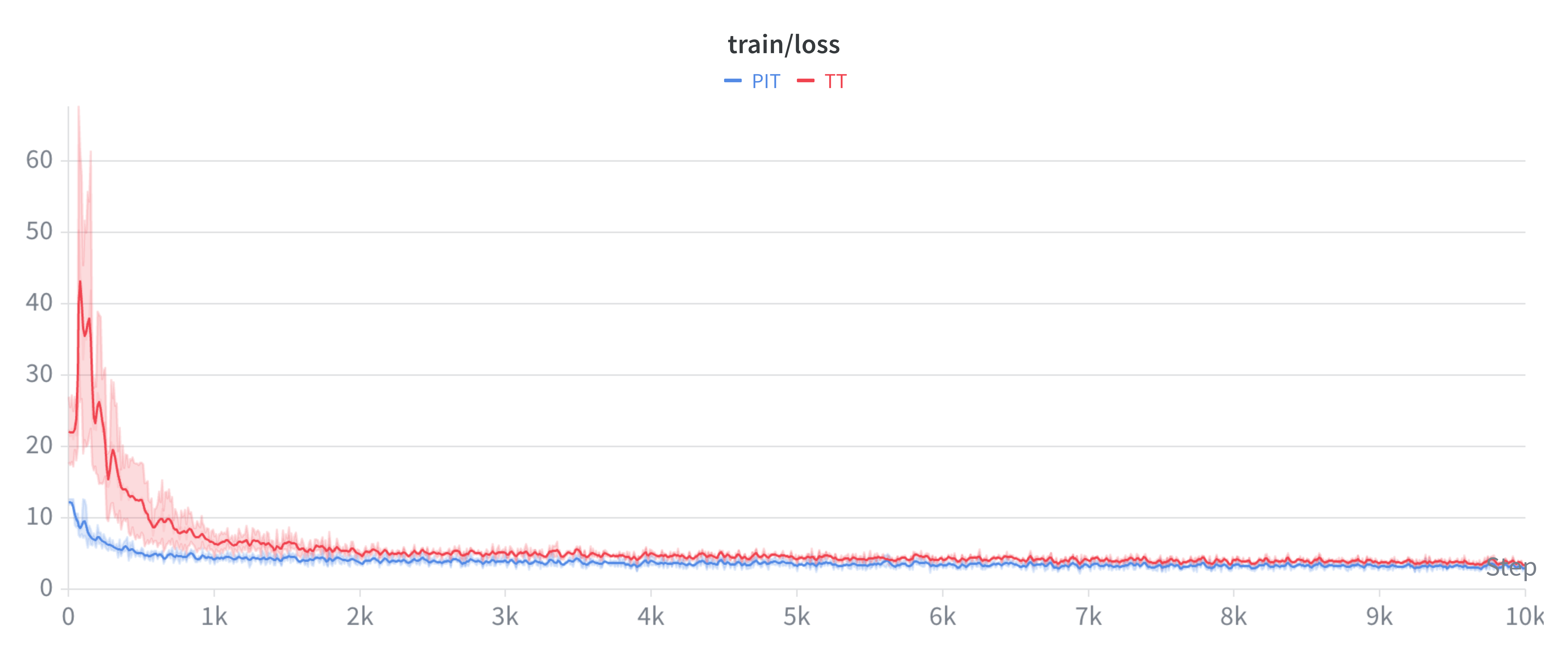}
  \caption{Representative continued-pretraining loss curves. PIT remains stable and reaches lower loss, while TT shows an early transient increase consistent with token-interface mismatch during the initial optimization phase.}
  \label{fig:loss_curve}
\end{figure}

We track training loss and perplexity as a function of tokens seen and compare optimization speed at fixed token budgets.
Figure~\ref{fig:loss_curve} shows representative pretraining loss curves on FineWeb-Edu 10B over the first 10k steps.
The comparison highlights the early continued-pretraining regime: PIT is stable from the beginning, while TT exhibits a transient loss increase before recovering.
This behavior is consistent with an interface mismatch being amplified by large early gradients when a checkpoint is exposed to a new corpus.

\paragraph{Analysis}
The loss curves suggest that PIT primarily improves the conditioning of the token interface rather than merely adding capacity.
Because the embedding and head are algebraically coupled, updates to hidden states are decoded through a synchronized vocabulary geometry throughout training.
This is most visible at the start of continued pretraining, when the model must adapt to new data while preserving useful token semantics from the source checkpoint.
The curve-level evidence complements the final losses in Table~\ref{tab:continued}: PIT not only ends lower, but also avoids the unstable early behavior observed under TT.

\subsection{Results on Token-Interface Consistency}

We next measure interface consistency directly.
Given an embedding $E$ and an unembedding $W_{\text{out}}$, a natural scalar diagnostic is the deviation from a left-inverse condition
\begin{equation}
\Delta_{\text{TI}} \;=\; \lVert W_{\text{out}}E - I_d\rVert_F,
\end{equation}
where smaller is better and $0$ indicates a perfectly consistent token interface.

Let $B_{\text{in}},B_{\text{out}}\in\mathbb{R}^{V\times d}$ denote input-side and output-side semantic bases.
For standard models, we extract them as the orthonormal factors from thin polar decompositions of the embedding matrix and the transposed unembedding matrix, computed in FP32.
For PIT, $B_{\text{in}}$ and $B_{\text{out}}$ coincide with the shared orthonormal memory $Z$ by construction up to numerical precision.
We compare $B_{\text{in}}$ and $B_{\text{out}}$ using three metrics, each with $0$ meaning perfect agreement: cosine distance, Procrustes error, and principal angle.

Table~\ref{tab:consistency_posttrain} reports input-side vs.\ output-side semantic alignment after continued pretraining on the diverse checkpoints.
PIT reduces cosine distance and Procrustes error to numerical zero and keeps principal angles near zero, while TT leaves measurable alignment errors.

\begin{table}[ht]
\centering

\caption{Token-interface consistency of weight tying on LM continued pretraining for diverse models.}
\label{tab:consistency_posttrain}

\resizebox{0.8\linewidth}{!}{

\begin{tabular}{llrrr}
  \toprule
  Model & Method & Cosine Distance $\downarrow$ & Procrustes Error $\downarrow$ & Principal Angle $\downarrow$ \\
  \midrule
  \multirow{2}{*}{\makecell{Granite4-350M-Base}} & TT & 0.0089 & 0.1333 & 0.0106 rad \\
                    & PIT & \tabh 0.0000 & \tabh 0.0000 & \tabh 0.0021 rad \\
  \midrule
  \multirow{2}{*}{\makecell{Granite4-350M-Instruct}} & TT & 0.0084 & 0.1293 & 0.0085 rad \\
                    & PIT & \tabh 0.0000 & \tabh 0.0000 & \tabh 0.0020 rad \\
  \midrule
  \multirow{2}{*}{\makecell{Qwen3-0.6B-Base}} & TT & 0.0165 & 0.1774 & 0.0611 rad \\
                    & PIT & \tabh 0.0000 & \tabh 0.0000 & \tabh 0.0024 rad \\
  \midrule
  \multirow{2}{*}{\makecell{Qwen3-0.6B-Instruct}} & TT & 0.0165 & 0.1778 & 0.0750 rad \\
                    & PIT & \tabh 0.0000 & \tabh 0.0000 & \tabh 0.0023 rad \\
  \bottomrule
  \end{tabular}
}
\end{table}

\paragraph{Analysis}
These diagnostics capture the mechanism that loss alone cannot isolate.
TT can still learn useful language-model behavior, but its input-side and output-side semantic bases need not remain aligned; the Qwen3 checkpoints show larger TT angles than the Granite checkpoints, suggesting that drift can be model-family dependent.
PIT removes this degree of freedom by construction, so the semantic bases used to encode and decode tokens remain synchronized throughout training.
This deterministic alignment explains why PIT is especially relevant for post-training updates and interpretability tools that rely on stable vocabulary directions: the same hidden-space displacement is less likely to be read through a changing output geometry.

\section{Discussion}
\label{subsec:discussion}

PIT can be viewed as a strict generalization of classic weight tying.
When the transform is the identity, PIT reduces to an orthonormal form of tying, with $E=Z$ and $W_{\mathrm{out}}=Z^\top$.
With a learned invertible transform, the embedding and unembedding paths become computationally asymmetric but remain pseudo-inverse consistent: $W_{\mathrm{out}}E=(TZ^\top)(ZT^{-1})=I_d$ when $Z^\top Z=I_d$.
Thus, $Z$ acts as a shared token memory, while $T$ specifies the hidden-space metric used to enter and leave that memory.

This construction gives compact language models an inductive bias beyond parameter sharing.
Standard transpose tying saves parameters, but it does not ensure that input-side and output-side token geometries remain synchronized during optimization.
PIT enforces this alignment by construction.
Our experiments reflect this distinction: in continued pretraining, PIT improves loss across diverse checkpoints and drives semantic-basis alignment errors to numerical precision, whereas TT leaves measurable interface drift.
The from-scratch results also clarify the boundary of the benefit: PIT is not always the strongest raw optimization bias from random initialization, but it yields a cleaner and more controllable interface once a model is being continued, adapted, or analyzed.

The main practical value is a more predictable token interface.
LoRA, patching, editing, and explainability probes often interpret hidden-space changes through the output vocabulary.
If the unembedding geometry has drifted from the input embedding, vocabulary-space effects become harder to anticipate.
PIT provides a stable bridge between hidden states and token probabilities, making token-level behavior easier to inspect and control.
This is especially relevant to logit-lens-style analysis and related vocabulary-space probes, where intermediate representations are projected through the output head.

PIT has modest overhead. It adds a hidden-size transform and triangular solves, while the dominant cost typically remains the vocabulary projection.
In practice, the important implementation choices are to keep the shared memory close to orthonormal, keep the transform invertible, and perform solves in FP32 when training in mixed precision.

\section{Conclusion}
\label{sec:conclusion}

We introduced \emph{Pseudo-Inverse Tying (PIT)}, a shared-memory parameterization that synchronizes a language model's input embedding and output projection through an orthonormal token memory and a learned invertible transform. PIT enforces a stable token interface by construction while avoiding explicit pseudo-inverse recomputation and vocabulary-sized auxiliary parameters.

Experiments show that PIT improves continued-pretraining stability and enforces near-exact input-output semantic alignment across settings. The results also reveal a useful boundary: from-scratch pretraining can favor the flexibility of conventional tying, while continued-pretraining and lightweight adaptation benefit from PIT's more predictable token geometry. Overall, PIT reframes weight tying as an interface-design problem, offering a practical route toward more stable and controllable compact language models.

% \begin{ack}
% Use unnumbered first level headings for the acknowledgments. All acknowledgments

% Do {\bf not} include this section in the anonymized submission, only in the final paper. You can use the \texttt{ack} environment provided in the style file to automatically hide this section in the anonymized submission.
% \end{ack}

\bibliographystyle{unsrt}
\bibliography{custom}

@article{gu2024vds,
	title   = {Vocabulary-Defined Semantics: Latent Space Clustering for Improving In-Context Learning},
	author  = {Gu, Jian and Aleti, Aldeida and Chen, Chunyang and Zhang, Hongyu},
	journal = {arXiv preprint arXiv:2401.16184},
	year    = {2024},
	url     = {https://arxiv.org/abs/2401.16184}
}

@inproceedings{gu2025semantic,
  title={A semantic-aware layer-freezing approach to computation-efficient fine-tuning of language models},
  author={Gu, Jian and Aleti, Aldeida and Chen, Chunyang and Zhang, Hongyu},
  booktitle={Findings of the Association for Computational Linguistics: ACL 2025},
  pages={8019--8033},
  year={2025}
}

@inproceedings{gu2025beyond,
  title={Beyond Neural Incompatibility: Cross-Scale Knowledge Transfer in Large Language Models through Latent Semantic Alignment},
  author={Gu, Jian and Aleti, Aldeida and Chen, Chunyang and Zhang, Hongyu},
  booktitle={Findings of the Association for Computational Linguistics: ACL 2026},
  year={2026}
}

@article{yang2025qwen3,
  title={Qwen3 technical report},
  author={Yang, An and Li, Anfeng and Yang, Baosong and Zhang, Beichen and Hui, Binyuan and Zheng, Bo and Yu, Bowen and Gao, Chang and Huang, Chengen and Lv, Chenxu and others},
  journal={arXiv preprint arXiv:2505.09388},
  year={2025}
}

@article{yenduri2023generative,
  title={Generative pre-trained transformer: A comprehensive review on enabling technologies, potential applications, emerging challenges, and future directions},
  author={Yenduri, Gokul and Srivastava, Gautam and Maddikunta, Praveen Kumar Reddy and Jhaveri, Rutvij H and Wang, Weizheng and Vasilakos, Athanasios V and Gadekallu, Thippa Reddy and others},
  journal={arXiv preprint arXiv:2305.10435},
  year={2023}
}

@misc{nostalgebraist2020logitlens,
  title        = {Interpreting GPT: The Logit Lens},
  author       = {nostalgebraist},
  year         = {2020},
  howpublished = {LessWrong post},
  url          = {https://www.lesswrong.com/posts/AcKRB8wDpdaN6v6ru/interpreting-gpt-the-logit-lens},
  note         = {Accessed 2026-01-16}
}

@article{elhage2021mathematical,
   title={A Mathematical Framework for Transformer Circuits},
   author={Elhage, Nelson and Nanda, Neel and Olsson, Catherine and Henighan, Tom and Joseph, Nicholas and Mann, Ben and Askell, Amanda and Bai, Yuntao and Chen, Anna and Conerly, Tom and DasSarma, Nova and Drain, Dawn and Ganguli, Deep and Hatfield-Dodds, Zac and Hernandez, Danny and Jones, Andy and Kernion, Jackson and Lovitt, Liane and Ndousse, Kamal and Amodei, Dario and Brown, Tom and Clark, Jack and Kaplan, Jared and McCandlish, Sam and Olah, Chris},
   year={2021},
   journal={Transformer Circuits Thread},
   note={https://transformer-circuits.pub/2021/framework/index.html}
}

@article{meng2022locating,
  title={Locating and editing factual associations in gpt},
  author={Meng, Kevin and Bau, David and Andonian, Alex and Belinkov, Yonatan},
  journal={Advances in neural information processing systems},
  volume={35},
  pages={17359--17372},
  year={2022}
}

@article{meng2022mass,
  title={Mass-editing memory in a transformer},
  author={Meng, Kevin and Sharma, Arnab Sen and Andonian, Alex and Belinkov, Yonatan and Bau, David},
  journal={arXiv preprint arXiv:2210.07229},
  year={2022}
}

@article{inan2016tying,
  title={Tying word vectors and word classifiers: A loss framework for language modeling},
  author={Inan, Hakan and Khosravi, Khashayar and Socher, Richard},
  journal={arXiv preprint arXiv:1611.01462},
  year={2016}
}

@inproceedings{press2017using,
  title={Using the output embedding to improve language models},
  author={Press, Ofir and Wolf, Lior},
  booktitle={Proceedings of the 15th Conference of the European Chapter of the Association for Computational Linguistics: Volume 2, Short Papers},
  pages={157--163},
  year={2017}
}

@book{absil2008optimization,
  title={Optimization algorithms on matrix manifolds},
  author={Absil, P-A and Mahony, Robert and Sepulchre, Rodolphe},
  year={2008},
  publisher={Princeton University Press}
}

@article{wen2013feasible,
  title={A feasible method for optimization with orthogonality constraints},
  author={Wen, Zaiwen and Yin, Wotao},
  journal={Mathematical Programming},
  volume={142},
  number={1},
  pages={397--434},
  year={2013},
  publisher={Springer}
}

@article{penedo2024fineweb,
  title={The fineweb datasets: Decanting the web for the finest text data at scale},
  author={Penedo, Guilherme and Kydl{\'\i}{\v{c}}ek, Hynek and Lozhkov, Anton and Mitchell, Margaret and Raffel, Colin A and Von Werra, Leandro and Wolf, Thomas and others},
  journal={Advances in Neural Information Processing Systems},
  volume={37},
  pages={30811--30849},
  year={2024}
}

@article{austin2021program,
  title={Program synthesis with large language models},
  author={Austin, Jacob and Odena, Augustus and Nye, Maxwell and Bosma, Maarten and Michalewski, Henryk and Dohan, David and Jiang, Ellen and Cai, Carrie and Terry, Michael and Le, Quoc and others},
  journal={arXiv preprint arXiv:2108.07732},
  year={2021}
}

@article{cobbe2021training,
  title={Training verifiers to solve math word problems},
  author={Cobbe, Karl and Kosaraju, Vineet and Bavarian, Mohammad and Chen, Mark and Jun, Heewoo and Kaiser, Lukasz and Plappert, Matthias and Tworek, Jerry and Hilton, Jacob and Nakano, Reiichiro and others},
  journal={arXiv preprint arXiv:2110.14168},
  year={2021}
}

@article{chen2021evaluating,
  title={Evaluating large language models trained on code},
  author={Chen, Mark},
  journal={arXiv preprint arXiv:2107.03374},
  year={2021}
}

@article{patel2021nlp,
  title={Are NLP models really able to solve simple math word problems?},
  author={Patel, Arkil and Bhattamishra, Satwik and Goyal, Navin},
  journal={arXiv preprint arXiv:2103.07191},
  year={2021}
}

@article{hu2022lora,
  title={Lora: Low-rank adaptation of large language models.},
  author={Hu, Edward J and Shen, Yelong and Wallis, Phillip and Allen-Zhu, Zeyuan and Li, Yuanzhi and Wang, Shean and Wang, Lu and Chen, Weizhu and others},
  journal={ICLR},
  volume={1},
  number={2},
  pages={3},
  year={2022}
}

@article{dey2023cerebras,
  title={Cerebras-gpt: Open compute-optimal language models trained on the cerebras wafer-scale cluster},
  author={Dey, Nolan and Gosal, Gurpreet and Khachane, Hemant and Marshall, William and Pathria, Ribhu and Tom, Marvin and Hestness, Joel and others},
  journal={arXiv preprint arXiv:2304.03208},
  year={2023}
}

@article{gao2026towards,
  title={Towards Intrinsic Interpretability of Large Language Models: A Survey of Design Principles and Architectures},
  author={Gao, Yutong and Meng, Qinglin and Zhou, Yuan and Pan, Liangming},
  journal={arXiv preprint arXiv:2604.16042},
  year={2026}
}

@misc{Soule_Bergmann_2025, title={IBM granite 4.0: Hyper-efficient, high performance hybrid models for Enterprise}, url={https://www.ibm.com/new/announcements/ibm-granite-4-0-hyper-efficient-high-performance-hybrid-models}, journal={IBM Granite 4.0: Hyper-efficient, High Performance Hybrid Models for Enterprise}, author={Soule, Kate and Bergmann, Dave}, year={2025}, month={Nov}}

%%%%%%%%%%%%%%%%%%%%%%%%%%%%%%%%%%%%%%%%%%%%%%%%%%%%%%%%%%%%

\appendix

% \section{Implementation Details}
% \label{appendix:details}

% \subsection{Environments}

% Our implementation uses deep learning framework \textsc{PyTorch}~\cite{Paszke2019PyTorchAI}, \textsc{Transformers}~\cite{Wolf2019TransformersSN}, and use \textsc{PEFT}~\footnote{https://github.com/huggingface/peft} to conduct the LoRA experiments.
% The LM finetuning experiments are based on existing PEFT methods, specifically \textsc{LoRA}~\cite{Hu2021LoRALA}.
% % In the experiments, we use zero-shot learning with prompt templates (see~\cref{tab:appendix_template}).
% We use quantization techniques (INT4) to load Llama3-8B, with the default settings~\cite{Dettmers2023QLoRAEF,Wu2023UnderstandingIQ}, which reduces the memory cost in LM finetuning with slight performance loss.

% The experiments are conducted via a single run, with the global random-seed $42$.
% The computation is based on a single Nvidia A100 (\num{80} \text{GB}), and the computation budget is around 500 GPU hours.

% \subsection{License and Terms}

% We understand and respect the licenses used in our experiments, including the Apache-2.0 license for Pythia models \footnote{https://github.com/EleutherAI/pythia} and the Llama3 community license \footnote{https://llama.meta.com/llama3/license/}.
% % 
% We confirm that our use of existing artifacts was consistent with their intended use.

\section{Technical Design and Details}
\label{appendix:design}

\subsection{Model Architectures and Scaling Targets}
\label{sec:architectures}

\begin{table}[ht]
  \centering
  \resizebox{0.6\linewidth}{!}{%
  \begin{tabular}{lrrrrrrr}
    \toprule
    Model & Params & Embed $\%$ & $L$ & $d_{\text{model}}$ & $n_h$ & $n_{kv}$ & $d_{\text{ff}}$ \\
    \midrule
    % \emph{S}mall   & 133 M & 18.5\% & 16 &  768 & 12 & 3 & 2304 \\
    Llama-0.3B  & 303 M & 10.8\% & 24 & 1024 & 16 & 4 & 2816 \\
    Llama-0.6B  & 613 M & 6.7\% & 32 & 1280 & 20 & 5 & 3584 \\
    Llama-1.2B  & 1.17 B & 4.4\% & 40 & 1600 & 24 & 6 & 4480 \\
    \bottomrule
    \end{tabular}  
   }
  \caption{Architectural targets for Llama-like models used in from-scratch pretraining, with vocabulary size $V{=}32{,}000$, gated GQA with $d_{\text{head}}{=}64$, and SwiGLU.}
\label{tab:sweetspots}
\end{table}

We evaluate three Llama-like decoder-only Transformer variants selected to align with compact on-device and edge-server parameter regimes while using a modern efficient block design. All models share:
(i) a fixed tokenizer with vocabulary size $V=32{,}000$,
(ii) tied input/output embeddings for the TT baseline or the PIT shared-memory parameterization,
(iii) grouped-query attention (GQA) with a gated attention output,
(iv) SwiGLU feed-forward networks,
and (v) a parallel Transformer block where the attention and MLP branches are computed in parallel and their outputs are combined in the residual stream.

\paragraph{Attention (Gated GQA)}
We fix head dimension $d_{\text{head}}=64$, hence $n_h = d_{\text{model}}/64$ query heads. We use GQA with
\begin{equation}
n_{kv} = \frac{n_h}{4}, \qquad d_{kv} = n_{kv}\, d_{\text{head}} = \frac{d_{\text{model}}}{4}.
\end{equation}
The attention output is modulated by a lightweight learnable channel-wise gate, whose parameter count is negligible relative to the projection matrices.

\paragraph{MLP (SwiGLU)}
We use SwiGLU with intermediate width
\begin{equation}
d_{\text{ff}} \approx \frac{8}{3} d_{\text{model}},
\end{equation}
rounded up to a kernel-friendly multiple of 256. This preserves the standard compute and parameter budget relative to a $4d_{\text{model}}$ GeLU MLP while benefiting from gating.

\paragraph{Parameter accounting}
With tied embeddings or PIT shared memory, the token-interface parameter budget is dominated by $Vd_{\text{model}}$ plus, for PIT, the additional hidden-space transform. Ignoring small terms such as biases and treating gating as negligible, the dominant per-layer parameters are:
\begin{equation}
P_{\text{attn}} \approx 2d^2 + 2d\, d_{kv}, \qquad
P_{\text{mlp}} \approx 3d\, d_{\text{ff}},
\end{equation}
with $d \equiv d_{\text{model}}$.
Reported parameter counts additionally include LayerNorm scale and shift parameters; positional embeddings are omitted for clarity.

\subsection{Implementation Details}

\paragraph{Shared memory $Z$}
PIT requires $Z^\top Z=I_d$.
For continued pretraining, we compute a thin polar decomposition of the source checkpoint embedding matrix and take the orthonormal factor as $Z$.
For from-scratch pretraining, we initialize $Z$ as a random Stiefel matrix, implemented by applying thin QR or polar decomposition to a random Gaussian matrix.
Unless otherwise noted, we freeze $Z$ after initialization, so interface adaptation happens through the learned SPD transform.
If training $Z$ is desired, we maintain Eq.~\eqref{eq:stiefel_Z} via polar retraction applied once per optimizer step:
\begin{equation}
Z \leftarrow \tilde{Z}\,(\tilde{Z}^\top \tilde{Z})^{-1/2},
\label{eq:Z_polar_retract}
\end{equation}
with all SPD operations performed in FP32 and an optional ridge $\epsilon I$ for numerical safety.

\paragraph{SPD transform $T$}
We parameterize $T\succ 0$ via a Cholesky factorization $T=LL^\top$ with positive diagonal, where the diagonal is mapped by softplus or $\exp(\cdot)$ to enforce positivity.
We learn $T$ through $L$ and compute triangular solves in FP32 for numerical stability.
To keep $T$ well-conditioned, we optionally clamp the diagonal of $L$ within a reasonable range; this is not required by Eq.~\eqref{eq:pinv_exact}, but it improves robustness in large-scale mixed-precision runs.
Optional regularizers on $\log\det T$ or on the diagonal scale of $L$ can be added, although our default setup relies on the SPD parameterization and standard optimization.

\paragraph{Efficient computation}
We compute logits as $(hT)Z^\top$, avoiding explicit formation of $W_{\text{out}}$.
For embedding lookup, we retrieve token rows $z_t$ from $Z$ and apply $T^{-1}$ via triangular solves instead of an explicit inverse.
This preserves the intended pseudo-inverse relationship while keeping overhead close to standard tying.

\paragraph{Checkpointing and deployment}
To stay consistent with standard weight-tying practice, we checkpoint $(Z,T)$ and reconstruct $E$ and $W_{\mathrm{out}}$ on load using Eq.~\eqref{eq:spt_def} when materialized matrices are needed.
This reduces redundant storage and keeps the tying relationship explicit.

\paragraph{Implementation decision ($m=d$)}
While PIT can be extended to a higher-dimensional shared memory, in this work we fix the shared-memory dimension to the hidden size for simplicity and fair comparison to standard language models:
(i) it keeps the embedding and head dimensionality unchanged;
(ii) it preserves the asymptotic cost of the vocabulary projection; and
(iii) it avoids introducing extra capacity that could confound the effect of pseudo-inverse tying.

\section{More Results}
\label{appendix:results}

\subsection{Weight Tying for LM From-Scratch Pretraining}

We pretrain on FineWeb-Edu~\cite{penedo2024fineweb} using its official 10B-token subset, which matches the data scale targeted by compact LMs.
We use checkpoint-native tokenizers in continued pretraining and a fixed tokenizer in from-scratch pretraining.
All pretraining runs use AdamW, a fixed context length of 1K tokens, and gradient accumulation to reach the same global batch size within each comparison.
We report final training loss, training perplexity, and runtime at fixed token budgets.

\begin{table}[ht]
\centering
\small
\caption{Effects of weight tying on from-scratch pretraining and continued pretraining for scaled models.}
\label{tab:scratch}

\resizebox{0.7\linewidth}{!}{
\begin{tabular}{cllrrr}
  \toprule
  Stage & Model & Method & Training Loss $\downarrow$ & Training PPL $\downarrow$ & Runtime $\downarrow$ \\
  \midrule
  \multirow{7}{*}{\rotatebox[origin=c]{90}{\makecell{From-Scratch\\Pretraining}}}
  & \multirow{2}{*}{\makecell{Llama-0.3B}} & TT & \tabh 2.875 & \tabh 17.722 & \tabh 1h 14m \\
                    && PIT & 3.915 & 50.166 & 1h 15m \\
  \cmidrule(lr){2-6}
  & \multirow{2}{*}{\makecell{Llama-0.6B}} & TT & \tabh 3.599 & \tabh 36.554 & \tabh 1h 50m \\
                    && PIT & 4.027 & 56.101 & 1h 51m \\
  \cmidrule(lr){2-6}
  & \multirow{2}{*}{\makecell{Llama-1.2B}} & TT & \tabh 2.907 & \tabh 18.303 & \tabh 3h 43m \\
                    && PIT & 4.174 & 64.984 & 3h 44m \\
  \midrule
  \multirow{7}{*}{\rotatebox[origin=c]{90}{\makecell{Continued\\Pretraining}}}
  & \multirow{2}{*}{\makecell{GPT-256M}} & TT & 2.311 & 10.080 & 1h 24m \\
                    && PIT & \tabh 2.304 & \tabh 10.015 & \tabh 1h 21m \\
  \cmidrule(lr){2-6}
  & \multirow{2}{*}{\makecell{GPT-590M}} & TT & \tabh 3.523 & \tabh 33.892 & 2h 03m \\
                    && PIT & 3.714 & 41.027 & \tabh 2h 00m \\
  \cmidrule(lr){2-6}
  & \multirow{2}{*}{\makecell{GPT-1.3B}} & TT & 3.025 & 20.593 & 3h 53m \\
                    && PIT & \tabh 2.821 & \tabh 16.789 & \tabh 3h 50m \\
  \bottomrule
  \end{tabular}
}

\end{table}

\paragraph{Analysis}
The scaled-model results show a regime-dependent trade-off.
For from-scratch pretraining, TT reaches lower loss and perplexity across all three Llama-like models, suggesting that PIT's strict interface constraint can be too restrictive during early representation formation from random initialization.
For continued pretraining, PIT improves GPT-256M and GPT-1.3B and slightly underperforms TT on GPT-590M, while maintaining similar or slightly lower runtime.
These results refine the main-message interpretation: PIT is most reliable when it regularizes an existing token interface during continued pretraining, whereas from-scratch optimization may benefit from the additional freedom of conventional tying before a stable semantic geometry has emerged.

\begin{table}[ht]
\centering

\caption{Token-interface consistency of weight tying on from-scratch pretraining and continued pretraining for scaled models.}
\label{tab:consistency_pretrain}

\resizebox{0.8\linewidth}{!}{
\begin{tabular}{cllrrr}
  \toprule
  Stage & Model & Method & Cosine Distance $\downarrow$ & Procrustes Error $\downarrow$ & Principal Angle $\downarrow$ \\
  \midrule
  \multirow{7}{*}{\rotatebox[origin=c]{90}{\makecell{From-Scratch\\Pretraining}}}
  & \multirow{2}{*}{\makecell{Llama-0.3B}} & TT & 0.1241 & 0.4925 & 0.1121 rad \\
                    && PIT & \tabh 0.0000 & \tabh 0.0000 & \tabh 0.0024 rad \\
  \cmidrule(lr){2-6}
  & \multirow{2}{*}{\makecell{Llama-0.6B}} & TT & 0.0810 & 0.3978 & 0.0808 rad \\
                    && PIT & \tabh 0.0000 & \tabh 0.0000 & \tabh 0.0025 rad \\
  \cmidrule(lr){2-6}
  & \multirow{2}{*}{\makecell{Llama-1.2B}} & TT & 0.0610 & 0.3459 & 0.0476 rad \\
                    && PIT & \tabh 0.0000 & \tabh 0.0000 & \tabh 0.0028 rad \\
  \midrule
  \multirow{7}{*}{\rotatebox[origin=c]{90}{\makecell{Continued\\Pretraining}}}
  & \multirow{2}{*}{\makecell{GPT-256M}} & TT & 0.0619 & 0.3497 & 0.0448 rad \\
                    && PIT & \tabh 0.0000 & \tabh 0.0000 & \tabh 0.0024 rad \\
  \cmidrule(lr){2-6}
  & \multirow{2}{*}{\makecell{GPT-590M}} & TT & 0.0315 & 0.2500 & 0.0344 rad \\
                    && PIT & \tabh 0.0000 & \tabh 0.0000 & \tabh 0.0025 rad \\
  \cmidrule(lr){2-6}
  & \multirow{2}{*}{\makecell{GPT-1.3B}} & TT & 0.0301 & 0.2444 & 0.0416 rad \\
                    && PIT & \tabh 0.0000 & \tabh 0.0000 & \tabh 0.0032 rad \\
  \bottomrule
  \end{tabular}
}
\end{table}

\paragraph{Results on Token-Interface Consistency}
Table~\ref{tab:consistency_pretrain} reports input-side vs.\ output-side semantic alignment for the scaled from-scratch and continued-pretraining models.
The alignment results are consistent with the main experiments: PIT keeps cosine and Procrustes errors at numerical zero and reduces principal angles to near-zero values, while TT exhibits nontrivial drift.
The from-scratch TT models have larger alignment errors than the continued-pretraining GPT models, indicating that unconstrained early optimization can produce stronger separation between input and output semantic bases even when it achieves lower raw training loss.
Thus, Table~\ref{tab:scratch} and Table~\ref{tab:consistency_pretrain} together show the key trade-off: PIT buys a stable token interface, but that stability is most beneficial when the training goal values controllability and continuation rather than maximum from-scratch optimization flexibility.

\subsection{Effects of Weight Tying on Downstream Adaptation}

We study lightweight adaptations using LoRA~\cite{hu2022lora}, where we fine-tune LMs on MBPP~\cite{austin2021program} for code and GSM8K~\cite{cobbe2021training} for math, then evaluate out-of-domain transfer on HumanEval~\cite{chen2021evaluating} for code and SVAMP~\cite{patel2021nlp} for math.
We report exact-match accuracy.

To keep the comparison controlled, we use the same LoRA configuration across methods at each scale.
The key hypothesis is not that PIT changes what the model can ultimately represent, but that a more consistent token interface makes small, low-rank updates easier to apply without inducing unintended shifts in how hidden states are read out as tokens.
This should matter most for transfer, where robustness to distribution shift is often limited by brittle reliance on a particular decoding geometry.

\begin{table}[ht]
\centering

\caption{Performance of weight tying on from-scratch and continued-pretraining checkpoints for downstream adaptation.}
\label{tab:performance}

\resizebox{0.7\linewidth}{!}{

\begin{tabular}{cllrrrr}
  \toprule
  Stage & Model & Method & MBPP $\uparrow$ & HumanEval $\uparrow$ & GSM8K $\uparrow$ & SVAMP $\uparrow$ \\
  \midrule
  \multirow{7}{*}{\rotatebox[origin=c]{90}{\makecell{From-Scratch\\Pretraining}}}
  & \multirow{2}{*}{\makecell{Llama-0.3B}} & TT  & \tabh 16.2 & \tabh 8.7 & \tabh 13.5 & \tabh 10.1 \\
                    && PIT & 15.6 & 8.2 & 13.1 & 9.6 \\
  \cmidrule(lr){2-7}
  & \multirow{2}{*}{\makecell{Llama-0.6B}} & TT  & \tabh 24.8 & \tabh 15.9 & \tabh 21.1 & \tabh 17.0 \\
                    && PIT & 24.2 & 15.4 & 20.6 & 16.6 \\
  \cmidrule(lr){2-7}
  & \multirow{2}{*}{\makecell{Llama-1.2B}} & TT  & \tabh 33.6 & \tabh 24.1 & \tabh 31.0 & \tabh 26.4 \\
                    && PIT & 32.9 & 23.6 & 30.4 & 25.8 \\
  \midrule
  \multirow{7}{*}{\rotatebox[origin=c]{90}{\makecell{Continued\\Pretraining}}}
  & \multirow{2}{*}{\makecell{GPT-256M}} & TT  & 18.7 & 10.3 & 16.9 & 12.4 \\
                    && PIT & \tabh 19.5 & \tabh 11.6 & \tabh 17.3 & \tabh 13.5 \\
  \cmidrule(lr){2-7}
  & \multirow{2}{*}{\makecell{GPT-590M}} & TT  & 28.9 & \tabh 19.6 & 25.7 & 20.8 \\
                    && PIT & \tabh 29.3 & 19.2 & \tabh 26.1 & \tabh 21.5 \\
  \cmidrule(lr){2-7}
  & \multirow{2}{*}{\makecell{GPT-1.3B}} & TT  & 39.5 & 29.8 & 36.2 & 31.1 \\
                    && PIT & \tabh 39.9 & \tabh 30.9 & \tabh 36.7 & \tabh 32.3 \\
  \bottomrule
  \end{tabular}
}

\end{table}

\paragraph{Analysis}
The adaptation results are mixed, and the pattern differs by training regime.
For continued-pretraining checkpoints, PIT is modestly better in most cells, with the largest gains appearing on harder transfer settings such as HumanEval and SVAMP.
This is consistent with the idea that stabilizing how hidden states map back to vocabulary items can make low-rank updates less brittle and reduce unintended shifts in decoding behavior.
For from-scratch checkpoints, TT is slightly better across the board in this set of runs.
A plausible explanation is that these checkpoints still benefit from the extra degrees of freedom implicit in transpose tying, while PIT's stricter interface constraint can mildly limit how the update reshapes the token readout.
Overall, PIT looks most useful as a robustness and controllability bias for continued training and lightweight updates on pretrained models, rather than as a universally dominant choice for from-scratch models.

\section{A Formal Description}

We summarize the training procedure for Pseudo-Inverse Tying (PIT) in Algorithm~\ref{alg:pit}. The method constructs a single shared token memory from either a source-checkpoint embedding for continued pretraining or a random orthonormal initialization for from-scratch pretraining. It then learns a compact SPD transform that jointly determines the input embedding map and the output projection used to produce logits. This implementation directly realizes the tied parameterization introduced in our equations, while avoiding explicit materialization of the embedding and unembedding matrices and preserving pseudo-inverse synchronization throughout training.

\begin{algorithm}
\caption{Pseudo-Inverse Tying (PIT) with Shared Memory $Z$ and SPD Transform $T$}
\label{alg:pit}
\begin{algorithmic}[1]
\REQUIRE Vocabulary size $V$, hidden size $d$, training corpus $\mathcal{D}$, mode $\mathtt{mode}\in\{\texttt{teacher},\texttt{scratch}\}$.
\REQUIRE (Optional) Teacher embedding table $E_0\in\mathbb{R}^{V\times d}$ if $\mathtt{mode}=\texttt{teacher}$.
\ENSURE Trained Transformer parameters $\Theta$, shared memory $Z\in\mathbb{R}^{V\times d}$, and SPD transform $T\in\mathbb{R}^{d\times d}$.

\vspace{2pt}
\STATE \textbf{Initialize shared memory $Z$ (column-orthonormal).}
\IF{$\mathtt{mode}=\texttt{teacher}$}
    \STATE Compute thin polar decomposition $E_0 = U H$ with $U^\top U = I_d$ and $H \succ 0$.
    \STATE Set $Z \leftarrow U$.
\ELSE
    \STATE Sample $\tilde{Z}\sim\mathcal{N}(0,1)^{V\times d}$ and compute thin QR factorization $\tilde{Z}=QR$.
    \STATE Set $Z \leftarrow Q$.
\ENDIF
\STATE Optionally freeze $Z$ for the remainder of training (recommended for controlled ablations).

\vspace{2pt}
\STATE \textbf{Parameterize the SPD transform $T$.}
\STATE Maintain a lower-triangular matrix $L\in\mathbb{R}^{d\times d}$ with positive diagonal (e.g., $\mathrm{diag}(L)=\exp(\ell)$).
\STATE Define $T \leftarrow LL^\top$.
\IF{$\mathtt{mode}=\texttt{teacher}$}
    \STATE (Optional) Initialize $L$ such that $T^{-1}$ approximately matches the teacher embedding scale induced by $H$.
\ELSE
    \STATE Initialize $L \leftarrow I_d$ (thus $T=I_d$).
\ENDIF

\vspace{2pt}
\STATE \textbf{Define the tied embedding/head maps (do not materialize full matrices).}
\STATE \textbf{Embed:} for token id $t$, compute $e_t \leftarrow z_t T^{-1}$, where $z_t$ is the $t$-th row of $Z$.
\STATE \textbf{Project:} for hidden state $h$, compute logits $\mathrm{logits}(h) \leftarrow (hT) Z^\top$.
\STATE \COMMENT{When $Z^\top Z=I_d$, these maps satisfy the pseudo-inverse consistency implied by the PIT equations.}

\vspace{2pt}
\STATE \textbf{Train with standard language-model objective.}
\FOR{each optimizer step}
    \STATE Sample minibatch $\mathcal{B}\subset\mathcal{D}$ and run the Transformer to obtain hidden states $h$ under parameters $\Theta$.
    \STATE Compute logits using PIT projection: $\mathrm{logits} \leftarrow (hT)Z^\top$.
    \STATE Compute loss $\mathcal{L}$ (e.g., next-token cross-entropy) and update $(\Theta, L)$ by backpropagation.
    \IF{$Z$ is trainable}
        \STATE Apply a Stiefel retraction to $Z$ to maintain $Z^\top Z \approx I_d$
        \STATE \COMMENT{e.g., polar retraction $Z\leftarrow Z(Z^\top Z+\epsilon I)^{-1/2}$, executed per optimizer step or periodically.}
    \ENDIF
\ENDFOR

\vspace{2pt}
\STATE \textbf{Return} $(\Theta, Z, T)$ with $T=LL^\top$.
\end{algorithmic}
\end{algorithm}

% \input{sections/appendix_d}

%%%%%%%%%%%%%%%%%%%%%%%%%%%%%%%%%%%%%%%%%%%%%%%%%%%%%%%%%%%%

% \newpage
% \input{checklist.tex}

\end{document}